# Formalization of Dialogue in the Decision Support System of Dr. Watson Type.

Authors: Saveli I Goldberg(1), Vladimir Sluchak

((1) MGH Radiation Oncology Department)

## Abstract

*The article further develops and formalizes a theory of friendly dialogue in an AI System of Dr. Watson type, as proposed in our previous publication[4],[19]. The main principle of this type of AI is to guide the user toward a solution in a friendly manner, using questions based on the analysis of user input and data collected in the system.*

## 1 Introduction

The use of AI systems [17] may run into a problem of distrust from specialists and loss of professional skills caused by the transfer of decision-making responsibility. Hence, the search for Friendly AI - a term coined by Eliezer Yudkowsky [18]. Dr.Watson-type of friendly AI was proposed in our previous publications [4],[19] and here we further develop its theory.

## 2 Principles of Dr. Watson-type systems (DWS).

When creating requirements for our friendly Decision Support System[19], we found a model of AI–human collaboration in the classic literary interaction between Mr. Holmes and Dr. Watson from the stories by Arthur Conan Doyle[9]. The principles of DWS reflect the qualities of Dr. Watson in dialogue with Mr. Holmes:
1. Dr. Watson is Mr. Holmes' chronicler. He documents Mr. Holmes' victories, failures, and reasoning.
2. Dr. Watson asks questions every time he misses or doesn't understand the logic in Mr. Holmes's train of thought, thus stimulating thinking and promoting Holmes's creativity.
3. Dr. Watson does not compete but rather collaborates with Mr. Holms. Therefore, Mr. Holmes doesn't perceive him as a threat.
4. Dr. Watson covers Mr. Holmes in dangerous situations, but he does not take over the cases even under those circumstances.
5. Mr. Holmes trusts Dr. Watson and does not expect him to create problems.
6. Collaboration between Mr. Holmes and Dr. Watson resembles a game in which both enthusiastically play their roles.

Translating these features of Dr. Watson's style and adding some technical items such as visualization, a user-friendly interface, and contextual directories we get the requirements for our system:
  1. The system should analyze and document the progress of the professional growth of the user. For example, the variability in the accuracy of their decisions/predictions from patient to patient, how the accuracy of their decisions/predictions changes over time, and how their performance compares with the average results in the specialty and/or a certain subgroup of specialists.
  2. The system should discover contradictions and omissions in the user's reasoning and help the user overcome these logical inconsistencies

3. The system-user interaction process should start with setting the goals for the desired outcome and the user's care plan. Subsequently, the system should influence decision-making by organizing and presenting the data or asking clarifying questions.
4. The system should provide alarms in situations of "obvious" errors – hence it should have a list of those errors.
5. The confidentiality of specialist-machine interaction must be ensured. The documentation of the user-system interaction and interim solutions should be unavailable to anyone except for the system.
6. The system should provide context-based references and help. Game elements should be included in the interaction between the user and the system.
7. The system should have a convenient, efficient, and friendly user interface

Evaluating and refining user solutions is fundamental for DWS. This implies the generation of internal DWS solutions at certain steps of the user-machine interaction. If a DWS-generated solution doesn't match the one by the user, DWS attempts to influence the user to reconsider. This is done by presenting questions generated by DWS at each step of the decision-making process.

Essentially DWS decision-making process is a *dialogue* that should lead to an optimal solution. It can be formalized as branching into two sub-processes running in parallel and exchanging information at certain points of their trajectories. The exchange of information is organized in a way that facilitates fast conversion of the process. This technique has proven to be effective and is widely used in machine computations [14]. In our case, one sub-process runs inside a machine and another one in a human's mind. The sides are not equally open at this exchange, DWS does not show its internal solutions; to make the information psychologically palatable the system presents it to the user implicitly – with questions that should stimulate a fast-thinking ("System1") process in the user's mind [10], and this feature should contribute to the effectiveness of the system.

Implicit, intuitive thinking is not commonly attributed to a machine, but the characteristics of this type of fast thinking (associativity, randomness, patterning, chaining, layering, networking, etc.) are used in DWS when analyzing user input and generating questions [11].

Yet another very important feature of the DWS is the ability to store the history of previous decisions and errors, analyze and generalize the causes of errors, and effectively present this data to the user at any step of the decision-making process. While presenting the history of previous decisions and errors DWS requires the user to assess the prognosis as a result of the selected actions. If the reality differs from the prognosis, the user is given the option to explain the discrepancy. The necessity of explaining the failure has a positive effect on the quality of the decisions in the long run.

The DWS process Shema (Fig 1) reflects the main states, actions, and transitions of the process, in particular, the branching into two subprocesses is seen at the first step, just after the parameters input and the very intentional "inequality" of information exchange is reflected – in the box where DWS machine computes its internal solution hidden from the user.

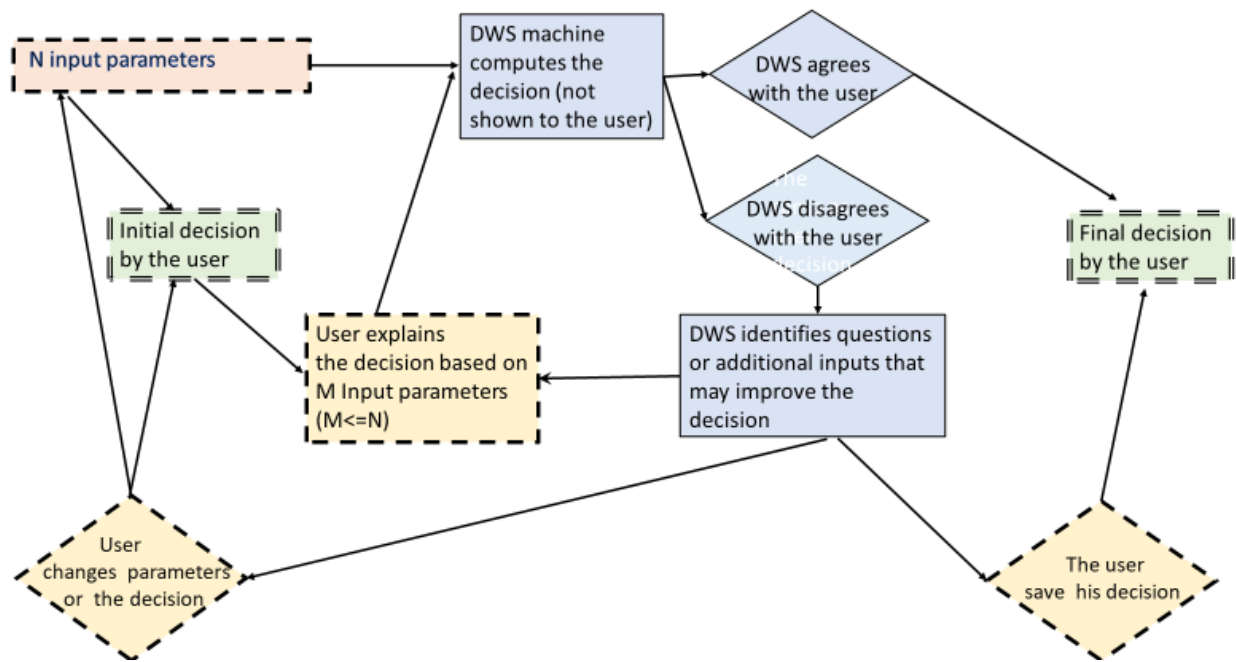

**Fig 1**. DWS process schema

Some learning systems use a so-called Socratic method of leading the user to a predefined solution with predefined questions organized as a tree and sequentially presented to the user at each step of the solution-finding process (for instance SASK [http://www.edutek.net/Kofi/sask/]). In DWS, the questions are generated "on the fly" in the user-machine dialog based on the input parameters and interim solutions provided by the user.

## 3 Formalization of user-DWS dialogue.

In the DWS process schema (Fig.1) the user-DWS dialogue is localized in the 5 central boxes.
Here are all the scenarios of that dialogue.

1. DWS shows the inconsistency between the user's decision and the parameters based on which he made the decision

2. DWS indicates the presence of additional information that may change the user's decision.

3. DWS shows that the input data could be distorted, and this could lead to a false decision.

4. DWS finds the precedents and asks the user to analyze them, the user can add precedents to the history.

We think this set of scenarios is complete, that is- it covers all cases of the decision-making process. This completeness is asserted by the last scenario, which is supposed to catch additional problems and precedents not stored and not discoverable by the machine algorithm but coming directly from the user.

Consider a formal description of the DWS-user dialogue:

*Let* **x** = [$x_1, x_2, ..., x_n$] be a ***vector*** of **n** parameters (***symptoms***, components) where $\mathbf{x_i}$ is a name-value pair and the value can be continuous (Numerical), categorical (Boolean), or descriptive (any type allowing to extract conditions and define mathematical operations on data).

*Let* **X** be the space of all such vectors, then **n** will be the dimension of it: **dim(X) = n.**

We assume that the values of vector components can always be presented as intervals or sets, so the methods of comparison will be well-defined. For instance, for the vectors $\mathbf{x}^1$ and $\mathbf{x}^2$, the conditions of equality of their components $\mathbf{x^1_i}$ and $\mathbf{x^2_j}$ with values given as ***numerical intervals*** will be defined as name($\mathbf{x^1_i}$) = name($\mathbf{x^2_j}$) with ***intersecting*** intervals of values.

*Let* vector $\mathbf{x}^2$ be defined as a sub-vector of $\mathbf{x}^1$ if each component of $\mathbf{x}^2$ is found in $\mathbf{x}^1$.

As the DWS machine is to operate on the vectors in **X,** the latter should be a metric space, that is the distance $\delta(\mathbf{x}^1, \mathbf{x}^2)$ between any pair of vectors in **X** should be defined. We assume that the distance $\delta(\mathbf{x}^1, \mathbf{x}^2)$ and the notion of proximity are defined specifically for each particular subject domain (see [Goldberg, Pinsky, 2022] and an example of the SAGE system further below).

*Let* $\Omega(x_i)$ be a set of values in the proximity of $x_i$ and $\Omega(x) \subset X$ - be the proximity of the vector x. For example, $\Omega(37.7^0 \text{ C})$ is ($37.4^0$ C – $38.0^0$ C), and $\Omega$(Small Headache) is (Small Headache, Moderate Headache).

*Let* $A \subset X$ be a subset of **X.**

Any vector **y** which is a sub-vector of some $\mathbf{x} \in A$ - will be called a syndrome of **A**.
A vector **y** will be called an ***antisyndrome*** of **A** if for every $\mathbf{x} \in A$ **y** is not a sub-vector of **x.**
Minimal antisyndrome of **A** is such an antisyndrome, that any sub-vector of it will be a syndrome of **A**.
For example, of all emergency calls, 35% were for men and 10% for pregnant patients. Combination {pregnant patient: True; gender: Male} will be a minimal antisyndrome for the set of ambulance patients.
A combination of "General aches" and *"*Headache" is minimal antisyndrome for "Airborne Allergy"

The ***minimal antisyndrome*** (Goldberg, S., 1984) of **A** plus any other parameter is an antisyndrome of **A**. If we exclude any parameter (symptom) from the minimal antisyndrome, the combination of remaining parameters becomes a syndrome for **A**. A complete set of minimal antisyndromes of **A** defines the boundary of this subset. (Minimal antisyndromes of **A** can be obtained from the experts or the training sample $A^* \subset A$.)

*Let* $\{\alpha_i\}$ be the set of possible solutions (set of diagnosis, for instance).

*Let* f(**x**): **X** -> $\{\alpha_i\}$ be a decision-making function (for example, specific diagnostic rules that each physician uses for illness definition).

*Let's* name $A(\alpha_i, f) \subset X$ a ***class*** of $\alpha_j$ if **f( x)**=$\alpha_i$**.**

Each solution α allows a typical representation (typical vector) s(α)=[$s_1(\alpha), s_2(\alpha), ...., s_n(\alpha)$] $\in$ **X**, which is either the center of gravity of the training sample of **A**(α,f) or a set of median values of parameters in such a sample. A typical vector for solution α can be provided by the experts, whose participation in the process is usually expected. For example, the typical picture of "Flu" is : {temperature > $38^0$ C, strong headache, severe general aches, weakness, extreme exhaustion, no cough, no stuffy running nose, and no sneezing}.

The interaction between the DWS machine and the user is a dialogue, where DWS poses questions, and the user responds. Before the start of this dialogue, the user provides his solution $\alpha_U$ and a vector of parameters $\mathbf{v} = [v_1,..v_m]$ (m<=n, where n is a number of all input parameters) based on which he came to his solution ($f_{user}(\mathbf{v})= \alpha_U$).

DWS machine has $f_{DWS}(\mathbf{x})$ for any $\mathbf{x} \epsilon \mathbf{X}$, a set of minimal antisyndromes of $\mathbf{A}(\alpha_j, f_{DWS})$ for expected solutions $\{\alpha_i\}$ and a set of typical representations of parameters as $\mathbf{s}(\alpha_j) = [s_1(\alpha_j), s_2(\alpha_j), ...., s_n(\alpha_j,)]$ for any $\alpha_j \epsilon \{\alpha_i\}$.

Basic scenarios for a dialogue between the DWS machine and the user are described in detail below. During the dialogue, the user can change both input data and decisions. Scenarios may change accordingly. The user solution will be denoted $\alpha_U$, DWS machine solution will be denoted $\alpha_W$. Algorithms for local explanation [13] play an important role in our process, especially those that assign weights of importance to the parameters for a given solution. For example, this could be LIME Local Interpretable Model-Agnostic Explanations [15] or SHAP Shapley Additive Explanations [12]. Below we consider the formal description of all scenarios along with an example of a case.

**DWS shows the inconsistency between the user's decision and the parameters based on which he made the decision**

It is assumed that the dialogue begins with the announcement of the user decision ($\alpha_U$) and vector of m<=n parameters ($\mathbf{v}$), based on which this decision was made. For example, the user describes his condition as slight general aches, sneezing, small headache, stuffy running nose, No Cough, Airborne Allergy in Anamneses, and his diagnosis is a relapse of Airborne Allergy. DWS machine checks if this set of options matches the selected solution. For this, the presence of antisyndromes of $\mathbf{A}(\alpha_U, f_{user})$ is checked. If antisyndromes are found, they are presented to the user for analysis (Fig 2). In our case "slight General aches" and *"small* Headache" is a minimal antisyndrome for "Airborne Allergy", so the user changes his diagnosis to Cold and DWS does not find the inconsistency.

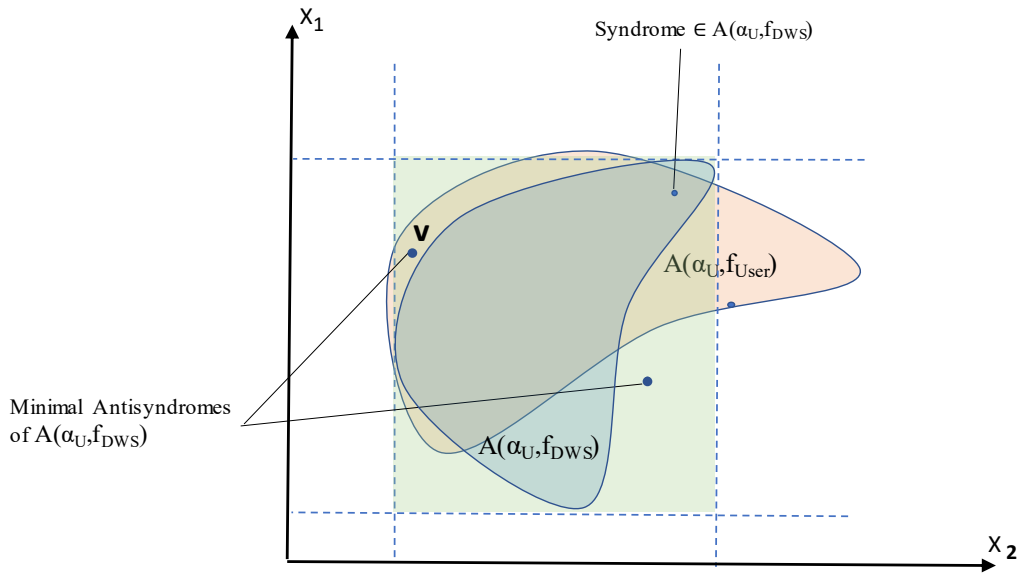

**Fig 2**. Example of the domains for the user solution $α_U$ in the space of 2 symptoms $(x_1,x_2)$ with numeric values. Each point of this space presents a syndrome or an antisyndrome. A $(α_U, f_{user})$ is the domain defined by the user, and A $(α_U, f_{DWS})$ is the domain defined by DWS for the same solution $α_U$. A $(α_U, f_{DWS})$ is bound by the area of minimal antisyndromes (green), and A $(α_U, f_{user})$ in our case overlaps this area. For each vector **v**$(x_1,x_2)$, that gets into the area of antisyndromes DWS machine will show the inconsistencies between **v** and $α_U$.

**DWS indicates the presence of additional information that may change the user's decision**.

Let the information in the sub-vector **v** from m parameters (m<n) be consistent with the user solution, in our case in the previous step the user changed the diagnosis from "Airborne Allergy" to "Cold".

However, there may be additional information that could make the user change his mind. DWS machine tries to check it and if such information exists then DWS asks the user to check it. DWS machine finds all vectors **vs**$_D$ = [**v**, $s_{m+1}(α_D),…, s_n(α_D)$] for all $α_D ∈ \{α_j\}$, where **s**$(α_W,)$ is a typical representation for $α_W$. If a case is found where $f_{DWS}$ (**vs**$_D$) = $α_D$ and $α_D ≠ α_U$, a local explanation of the **vs**$_D$ decision is sought (Fig 3).

High fever and extreme exhaustion should change the diagnosis from cold to flu. Based on local explanations, looking at the weights of parameters DWS machine finds the most important of those not provided by the user (not included in vector **v**) and poses a question about the importance of this parameter.

In our example, the first question is about fever. After responding to this question, the user can insist on his decision or change it. As 37.8 is not a high fever the user saves the diagnosis "Cold". The next question from DWS will be about Extreme exhaustion. Let the answer is "No", then the diagnosis of Cold remains.

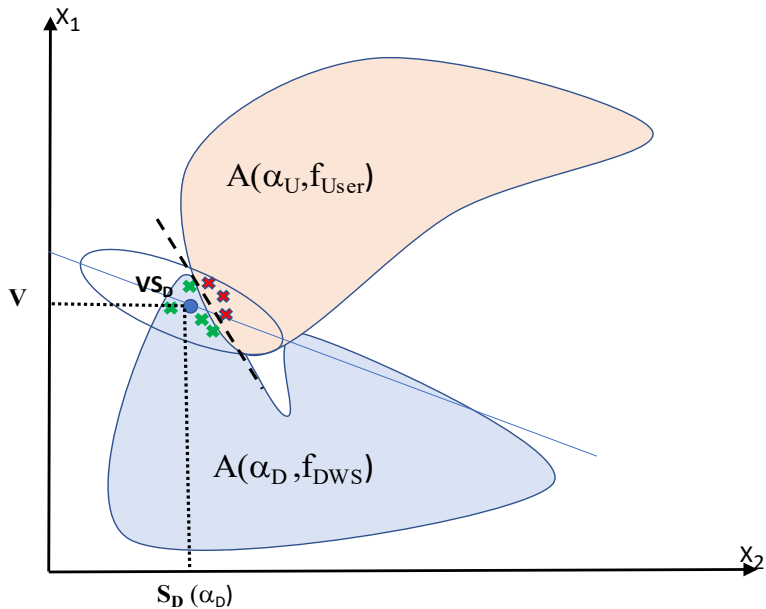

**Fig 3** Domains of solutions in the space of 2 variable vectors of parameters: initial vector $x_1$ with **m** parameters and additional vector $x_2$ with **n-m** parameters. DWS machine checks whether additional parameters can change the user's solution. If yes DWS machine uses a local explanation method (for instance LIME) to generate questions on the importance of parameters.

**DWS shows that the input data could be distorted, and this could lead to a false decision.**

Let the data be consistent with the user's decision and there are no additional parameters that should influence this decision. However, information can be distorted by measurement errors. The system checks whether such distortions can affect the user's decision, and if such a danger exists, the system asks to clarify the dangerous parameters. Fever could be 38.0, Headache could be Moderate, exhaustion could be small, and, in this situation, our case could be "Flu"

At this step, the DWS machine randomly selects vectors from the proximity **Ω(v)** of vector **v** ($f_{user}(v) = α_U$). Let there are vectors $x \in X$, where $f_{DWS}(x) = α_D$ and the solution $α_D \neq α_U$ and $x \in Ω(v)$. Let's select $x_{DWS}$ which has a minimum value of the metric $δ(x,v)$. A local explanation for $x_{DWS}$ is sought by the DWS machine (Fig. 4). The question on the meaning of the most sensitive parameters in this explanation is asked. Such a question may concern the refinement of parameters already present in **v** or not yet involved in the description of the case. DWS asks to check the temperature again. The user gets 38.0 and begins to think about "Flu".

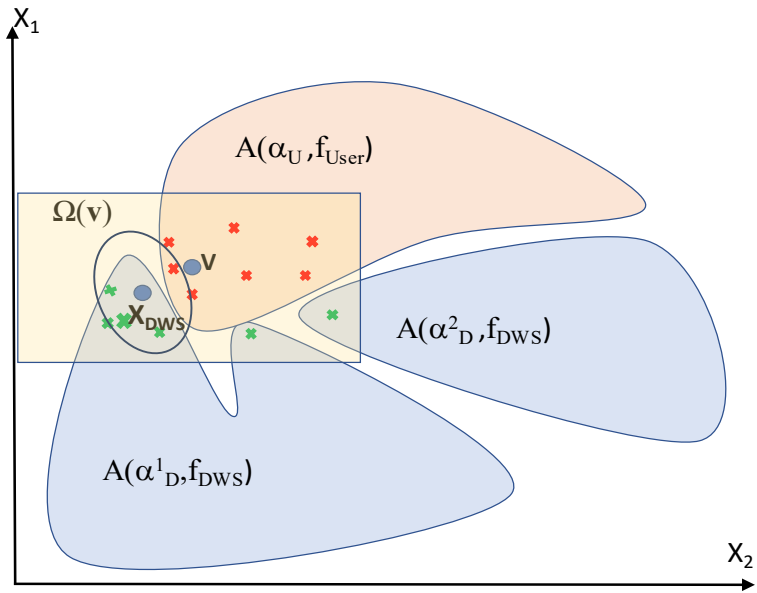

**Fig 4** Domains of solutions in the space of 2 variable parameters with numeric values. DWS machine checks the stability of the user solution with respect to the measurement errors in vector **v**. DWS finds the closest point $x_{DWS}$ with a different decision and then uses a local explanation procedure for $x_{DWS}$. The result of this local explanation is presented to the user.

**DWS finds precedents and asks the user to analyze them**.

The precedents are collected in the DWS: for each case, the user is asked to explain the errors and predict the consequences of their decision. As for the prediction, if the reality differs, the user has to explain the discrepancy. The need to predict the results and explain the failures has a positive effect on the quality of solutions in the long term. The explanations are collected and analyzed by the DWS, and the analysis is offered to the user in the following forms

1. Summary table of error descriptions.

2. Warnings about possible errors. The system selects the most probable error in each situation and shows it with an explanation of the probable cause. Since the current situation will not be identical to the one found in the history, the user is presented with a table of explanations for his error, sorted by the proximity of the situations. The computation of proximity [7] is based on

- Measurement errors

- Evaluation of parameters according to the Norm-Pathology Scale

-Difference in the informational weight of the parameters

Users can filter and sort the entries in the Error Explanation Table according to their preferences. The result of it in our case could be "Check the fever every 2 hours in the first 2 days". Sometimes the requirement to describe the causes of errors may seem not safe to the user, so DWS must ensure that such information does not become available to third parties. Also, the system can allow the user to edit the description of the cause of the error, and only the latest update is stored in the history. The loss of information in this case is offset by the effect of a double analysis of one's mistakes.